\documentclass[sigconf]{acmart}

\setcopyright{cc}
\setcctype{by}
\acmConference[MM '26]{Proceedings of the 34th ACM International Conference on Multimedia}{November 10--14, 2026}{Rio de Janeiro, Brazil.}
\acmBooktitle{Proceedings of the 34th ACM International Conference on Multimedia (MM '26), November 10--14, 2026, Rio de Janeiro, Brazil}
\acmYear{2026}
\copyrightyear{2026}
\acmDOI{10.1145/3767308.3835995}
\acmISBN{979-8-4007-2213-4/2026/11}

\usepackage{makecell}
\usepackage{enumitem}

\newcommand{\best}[1]{\textbf{#1}}

\title{R4DSG: Relative 4D Scene Graph Memory for Object-Centric Question Answering in Long Egocentric Video}

\author{Ke Ma}
\orcid{0000-0002-2085-5028}
\affiliation{%
  \department{College of Design and Innovation}
  \institution{Tongji University}
  \city{Shanghai}
  \country{China}}
\email{make@hust.edu.cn}

\author{Yamin Mao}
\orcid{0009-0003-2768-8722}
\affiliation{%
  \institution{Samsung R\&D Institute China - Beijing}
  \city{Beijing}
  \country{China}}
\email{yamin18.mao@samsung.com}

\author{Weiming Li}
\orcid{0000-0003-4054-5956}
\affiliation{%
  \institution{Samsung R\&D Institute China - Beijing}
  \city{Beijing}
  \country{China}}
\email{weiming.li@samsung.com}

\author{Shuai Tan}
\orcid{0000-0003-3322-5161}
\affiliation{%
  \institution{Shanghai Jiao Tong University}
  \city{Shanghai}
  \country{China}}
\email{tanshuai0219@sjtu.edu.cn}

\author{Yijie Zhong}
\orcid{0000-0002-2351-8799}
\affiliation{%
  \department{College of Design and Innovation}
  \institution{Tongji University}
  \city{Shanghai}
  \country{China}}
\email{dun.haski@gmail.com}

\author{Hao Chen}
\orcid{0000-0003-0814-9144}
\affiliation{%
  \department{Samsung Networks}
  \institution{Samsung Research America}
  \city{Plano}
  \state{Texas}
  \country{United States}}
\email{hao.chen1@samsung.com}

\author{Haofen Wang}
\orcid{0000-0003-3018-3824}
\affiliation{%
  \department{College of Design and Innovation}
  \institution{Tongji University}
  \city{Shanghai}
  \country{China}}
\email{carter.whfcarter@gmail.com}

\author{Meng Wang}
\correspondingauthor
\orcid{0000-0002-2293-1709}
\affiliation{%
  \department{College of Design and Innovation}
  \institution{Tongji University}
  \city{Shanghai}
  \country{China}}
\affiliation{%
  \department{Shanghai Research Institute for Intelligent Autonomous Systems}
  \institution{Tongji University}
  \city{Shanghai}
  \country{China}}
\email{mengwangtj@tongji.edu.cn}

\renewcommand{\shortauthors}{Ke Ma et al.}

\begin{document}

\begin{abstract}
Long-horizon egocentric video is a rich substrate for wearable AI assistants, but object-centric questions such as where an item was moved, when it last changed state, or why it was relocated remain difficult because caption- and transcript-based memories rarely preserve persistent object identity or structured spatial change. Existing long-video QA methods mainly emphasize temporal grounding and clip retrieval, while prior 3D scene-graph methods typically assume stronger geometry than free-motion wearable RGB video provides, including point clouds, RGB-D input, posed views, sparse reconstruction, or reconstructed scenes. R4DSG introduces a \textbf{relative 4D scene graph memory} for long egocentric video. Instead of storing raw graph sequences, R4DSG converts video into compact queryable memory entries indexed by time, place, persistent objects, anchor-relative change, and local interaction context. The main idea is to separate stable anchors from dynamic objects, maintain persistent object identity across frames, and represent object state through anchor-relative transitions rather than a globally aligned world model. Built on recent RGB-only advances in promptable video segmentation, temporal propagation, and relative 3D lifting, the method produces a retrieval-ready memory directly usable for long-horizon question answering. Evaluation on a 255-question object-related subset from EgoLifeQA shows, under question-only retrieval, a 6.7-point overall gain over EgoRAG-Text and a 12.5-point gain on \texttt{when} questions, which highlights the value of temporally organized object memory. These results position relative 4D scene graphs as a practical memory substrate for wearable assistants, AR systems, and embodied multimedia agents. GitHub Page: \url{https://dualtransparency.github.io/R4DSG/}.
\end{abstract}


\maketitle

\begin{figure*}[t]
  \centering
  \includegraphics[width=\textwidth]{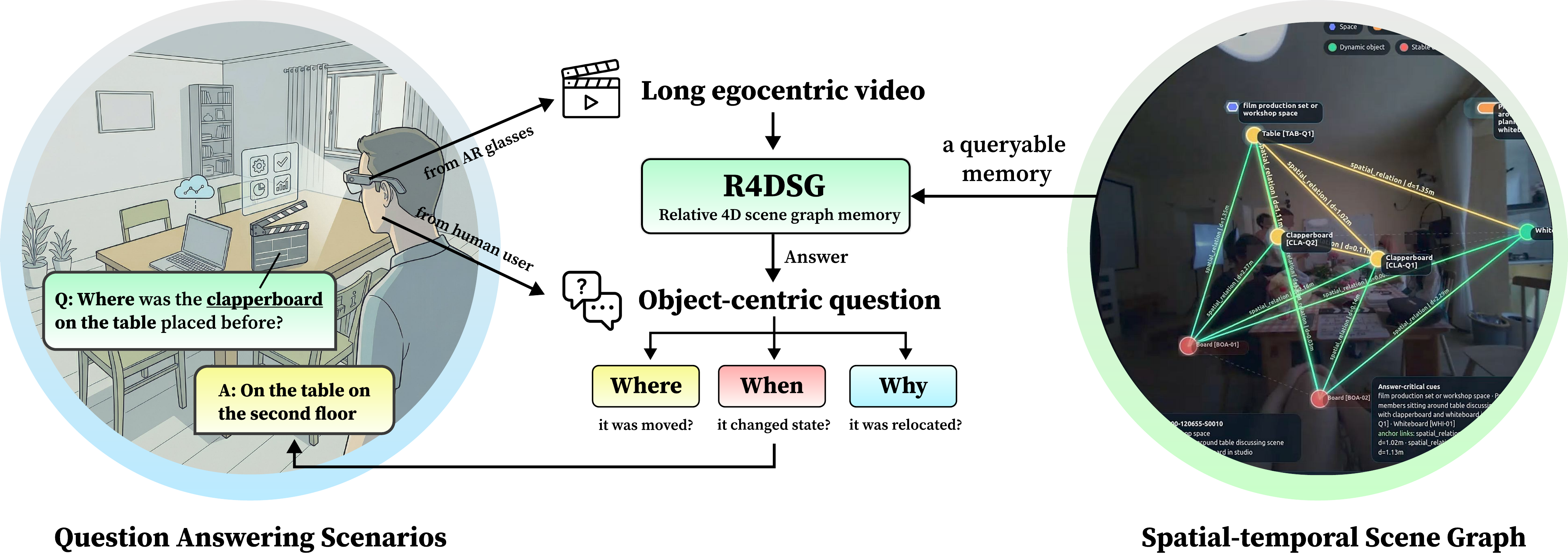}
  \caption{R4DSG maps long egocentric video to a relative scene-memory for object-centric question answering.}
  \Description{An overview showing a user asking an object-centric question over long egocentric video, the R4DSG memory module in the middle, and a spatial-temporal scene graph view on the right.}
  \label{fig:overview}
\end{figure*}

\section{Introduction}

Wearable cameras and AI glasses are moving multimedia systems from passive recognition toward persistent assistance. In that setting, a useful assistant is not limited to describing the current view. It must also answer questions grounded in past interactions with objects and spaces, such as \emph{Where did I leave the charger?}, \emph{When did I last move the mug from the desk?}, or \emph{Why did I take the bowl out of the fridge?} Recent work on egocentric devices and wearable assistants makes this trajectory increasingly explicit, spanning egocentric multimodal sensing platforms, context-aware AR assistants, proactive smart-glasses systems, and life-assistant benchmarks~\cite{engel2023aria,lee2024gazepointar,cai2025aiget,pu2025promemassist,yang2025egolife,ma2019design,ma2026phys,ma2026transbiolab,shi2017digital,wang2024towards}. Figure~\ref{fig:overview} shows R4DSG.

The central challenge is the need for a \emph{queryable memory}. Object-centric questions depend on preserving object identity, the stable references surrounding the object, and the transitions that explain its current state. Flat summaries, captions, or retrieved clips are often enough for coarse event recall, but they are weak at preserving structured evidence for object-state change. A useful memory for egocentric assistance should therefore satisfy four properties. It should preserve persistent object identity across time, encode spatial relations rather than only local appearance, compress long streams into compact retrievable units, and expose those units in a form directly usable by downstream QA. 

Recent progress in long egocentric video understanding sharpens this gap but does not close it. Grounded Multi-Hop VideoQA shows that many questions depend on multiple scattered evidential moments rather than one contiguous segment~\cite{chen2025multihop}. AMEGO argues that very long egocentric streams should be converted into an explicit reusable memory instead of being processed end to end each time~\cite{goletto2024amego}. EgoLife extends this agenda to a life-assistant setting and highlights memory retrieval, event recall, habit understanding, and relation reasoning as core capabilities for future egocentric assistants~\cite{yang2025egolife}. Yet the dominant abstractions remain caption-centric or summary-centric. They record events in language, but they do not maintain a persistent object-level account of an object’s location, its relations to stable references, when those relations changed, and the local context that explains such changes.

A natural answer is to use scene graphs. At the representation level, scene graphs are appealing because they bind entities, relations, and evidence into compact structured units that can be updated, searched, and reused~\cite{wu2024graphqa,zhao2023hostsg,chu2025graphvideoagent}. Prior work has shown that 3D scene graphs provide compact and semantically rich representations of objects and relations~\cite{wu2021scenegraphfusion,wu2023incremental,zhang2024egosg}. Open3DSG, VL-SAT, OpenFunGraph, and 3DGraphQA further demonstrate that open-vocabulary semantics, language guidance, and graph reasoning can make structured 3D representations more expressive downstream~\cite{wang2023vl,koch2024open3dsg,zhang2025openfungraph,wu2024graphqa}. These results are highly relevant, but they still assume stronger geometry than the target setting here provides, including point clouds, RGB-D input, posed views, sparse reconstruction, or reconstructed indoor scenes. In daily egocentric capture, the camera moves with the wearer, observations are partial, and a globally aligned world coordinate system is often unavailable, unreliable, or simply unnecessary~\cite{engel2023aria,goesele2025imaging,zhang2024egosg}.

Recent RGB-only foundation models for open-set video segmentation and tracking, temporal propagation, and monocular 3D lifting make this route technically plausible under weaker sensing conditions~\cite{ravi2024sam2,carion2025sam,karaev2025cotracker3,wang2024dust3r,leroy2024grounding,chen2025sam}. In particular, SAM 3 strengthens promptable video segmentation with concept-aware temporal propagation, while SAM 3D, DUSt3R, and MASt3R provide RGB-only routes to relative 3D cues without requiring depth input or pre-existing point clouds. Together, these advances make it feasible to build a queryable memory that captures relative spatial change in long egocentric video. What matters instead is whether an object moved away from one stable reference and toward another, whether that transition persisted, when it occurred, and what local interaction context best explains it. 

Motivated by this view, we introduce R4DSG, a \textbf{relative 4D scene graph memory}, which converts long egocentric video into queryable memory entries indexed by time, place, persistent objects, anchor-relative change, and local interaction context. The key insight in R4DSG is that long-horizon object memory can be organized around \emph{static anchors} and \emph{dynamic objects}. Static anchors such as tables, shelves, counters, drawers, sofas, or fridges provide stable relational references. Dynamic objects accumulate state changes by forming new spatial relations to different anchors over time. Instead of forcing all observations into one global frame, the method associates object instances across frames, retains only anchor-relative transitions that remain stable over time, and promotes those transitions into memory entries. The result is a memory that is compact enough for retrieval yet structured for object-centric reasoning.

The main contributions are as follows:
\begin{itemize}[leftmargin=1.2em]
    \item A relative 4D scene-graph formulation for long egocentric RGB video that represents object state through persistent anchor-relative transitions rather than a globally aligned map.
    \item A queryable memory design that converts frame-level graph evidence into segment-level retrieval documents while preserving place, activity, object state, actor, interaction, and edge cues needed by long-horizon QA.
    \item A pipeline grounded in SAM3-style consistency and RGB-only 3D lifting, followed by persistent identity association, static-anchor inference, and retrieval-aware memory writing.
\end{itemize}

\section{Related Work}

\subsection{Long egocentric video understanding, memory, and question answering}

Egocentric video understanding has evolved from action recognition toward long-horizon memory and question answering. EMQA is an early formulation of episodic memory QA, where an egocentric agent answers grounded questions using an explicit scene memory~\cite{datta2022episodic}. EgoSchema later established very long-form multiple-choice QA as a diagnostic benchmark for temporal understanding in egocentric video~\cite{mangalam2023egoschema}. Building on the Ego4D ecosystem~\cite{grauman2022ego4d}, GroundVQA formalized grounded QA in long egocentric videos and coupled answer generation with temporal localization of relevant evidence~\cite{di2024groundvqa}. Grounded Multi-Hop VideoQA further increased the reasoning burden by requiring multiple discontiguous evidence segments for a single question~\cite{chen2025multihop}. AMEGO proposed an active-memory representation for very long egocentric video that captures recurring locations and object interactions without repeatedly processing the full stream~\cite{goletto2024amego}. EgoLife broadened the task to an egocentric life assistant and introduced long-context tasks around event recall, relation reasoning, habit understanding, and memory-intensive assistance~\cite{yang2025egolife}.

These works establish the importance of long-term memory in egocentric video, but their dominant abstractions are still clips, summaries, captions, or high-level semantic memories. They are effective for generic recall and temporal localization, yet they do not explicitly target persistent object-state memory grounded in relative spatial relations. This direction also aligns with graph-aware retrieval work, where G-Retriever, KG$^2$RAG, and GNN-RAG show that once knowledge is expressed as a graph, retrieval can operate over compact structured evidence rather than flat text chunks~\cite{he2024gretriever,zhu2025knowledge,mavromatis2025gnnrag}. R4DSG focuses on the earlier multimedia problem: constructing such a queryable graph memory directly from long egocentric video.

\subsection{3D scene graphs and open-vocabulary structured scene understanding}

Scene graphs provide an explicit representation of entities and their relations, which makes them appealing for reasoning-heavy visual tasks. In 3D perception, 3DSSG introduced a benchmark and a learning framework for semantic scene graphs from indoor 3D reconstructions~\cite{wald2020learning}. SceneGraphFusion then showed that 3D scene graphs can be predicted incrementally from RGB-D sequences, which enables online graph construction under partial observations~\cite{wu2021scenegraphfusion}. Wu et al. later demonstrated that consistent 3D semantic scene graphs can also be incrementally constructed from RGB sequences by coupling sparse mapping with graph prediction~\cite{wu2023incremental}. VL-SAT injected visual-linguistic supervision into 3D semantic scene graph prediction from point clouds~\cite{wang2023vl}. Open3DSG moved toward open-vocabulary 3D scene graphs from point clouds~\cite{koch2024open3dsg,pan2023find}. EgoSG brought 3D scene graphs closer to the egocentric setting by learning from unposed RGB-D sequences without relying on reconstruction algorithms or camera poses~\cite{zhang2024egosg,eccvw}. OpenFunGraph extended structured 3D perception to functional relationships and downstream interaction-oriented reasoning in real-world indoor scenes~\cite{zhang2025openfungraph,ma2026wpis}. In ACM MM, 3DGraphQA further demonstrated that explicit scene-graph reasoning can improve QA over 3D scenes by providing a more interpretable intermediate structure~\cite{wu2024graphqa}.


These methods strongly motivate graph-structured 3D reasoning, but they address settings that differ from ours in a critical respect. They typically rely on point clouds, RGB-D input, posed multi-view observations, sparse mapping, or reconstructed scenes, whereas our target input is long, free-motion monocular egocentric RGB video. Accordingly, our goal is not to recover a globally consistent 3D scene. Instead, we seek a relative 3D abstraction sufficient for long-horizon memory and object-centric reasoning, where pairwise distances, anchor-relative transitions, and persistent identities matter more than metric global alignment. This relative geometry links object observations over time and supports compact, retrieval-ready memory for downstream question answering.

\subsection{Temporal scene graphs and graph-based video reasoning}

A parallel line of work studies temporal or spatio-temporal graph representations for video reasoning. Cherian et al. introduced a $(2.5+1)$D spatio-temporal scene graph for video QA, combining pseudo-3D scene structure with temporal graph reasoning~\cite{cherian2022stsg}. Urooj et al. proposed situation hyper-graphs for video QA, explicitly modeling actors, objects, and relations over time~\cite{urooj2023situation}. In ACM MM, HostSG constructed a holistic spatio-temporal scene graph for video semantic role labeling by merging clip-level dynamic scene graphs while preserving static and dynamic cues~\cite{zhao2023hostsg}. More recently, Action Scene Graphs introduced temporally evolving graphs tailored to long-form egocentric video~\cite{rodin2024action,zhao2026resilphase}. GraphVideoAgent showed that entity-relation graphs can also guide frame selection and reasoning in long-form video understanding~\cite{chu2025graphvideoagent}. At the zero-shot end, SAMJAM combined segmentation-and-tracking with VLM semantics for egocentric kitchen video scene graph generation, which demonstrates that foundation-model pipelines can produce temporally consistent graphs without task-specific training~\cite{li2025samjam}.

These works show that graph structure is an effective language for long-video reasoning, but they still leave open the problem addressed here. Existing temporal graph methods are often action-centric or event-centric, and even the most relevant zero-shot egocentric graph formulations focus on graph generation in narrower domains rather than on building a long-horizon memory whose atomic unit is an anchor-relative object change with explicit timestamps and retrievable contextual evidence.

\section{Method}
\label{sec:method}

\subsection{Task formulation}

Given a long egocentric RGB video $X = \{I_t\}_{t=1}^{T}$, the goal is to build a queryable memory $M$ that supports object-centric long-context QA under weak sensing conditions. The input is monocular, map-free, and only partially observed. No depth, point clouds, calibrated poses, or globally aligned world coordinates are assumed. Instead, the memory should preserve persistent object identity and object-state change over time.

The problem is written as
\begin{equation}
M = f(X), \qquad (\hat{y}, \hat{e}) = g(M, q),
\end{equation}
where $q$ is a question, $\hat{y}$ is the predicted answer, and $\hat{e}$ is the retrieved evidence. The focus is on object-centric questions whose answers depend on past state changes rather than on the current frame alone, especially \texttt{where}, \texttt{when}, and \texttt{why} questions.

\begin{figure*}[t]
    \centering
    \includegraphics[width=\textwidth]{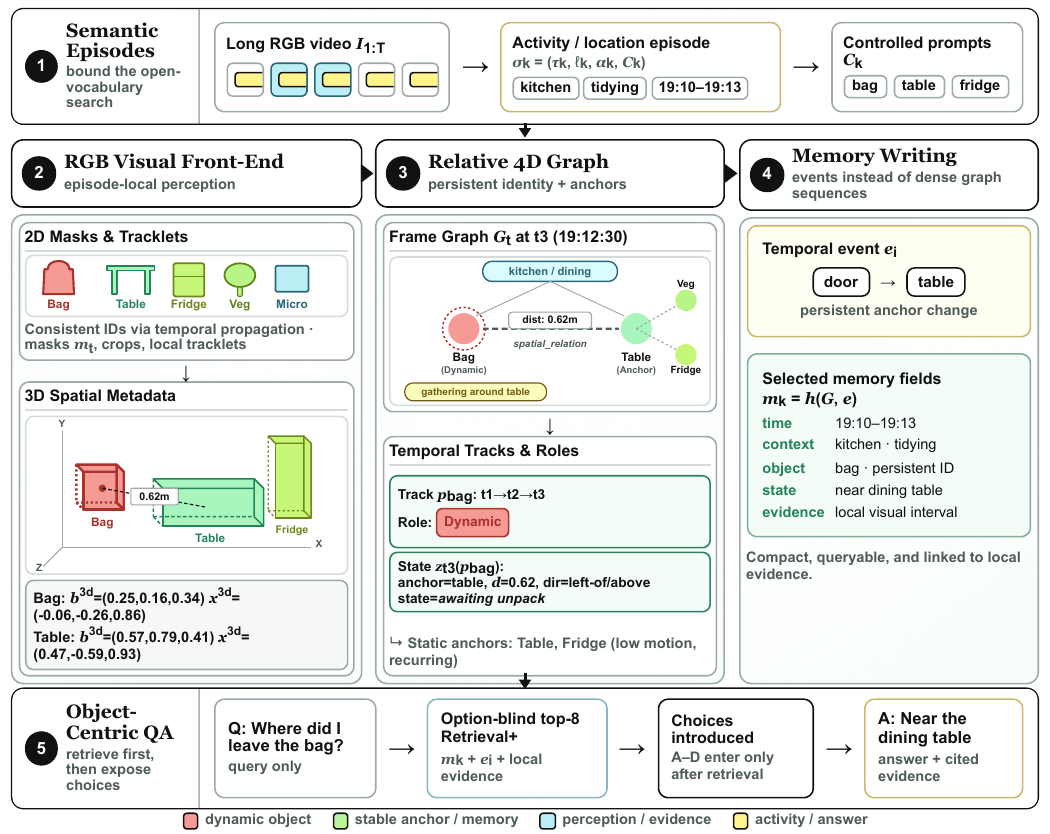}
    \caption{R4DSG pipeline. Semantic episodes guide RGB-only segmentation and lifting; persistent identities and anchor changes are then written as retrieval-ready memory for QA. The figure reads top-to-bottom, with its three core technical stages arranged left-to-right between episode parsing and downstream answer synthesis.}
    \Description{A five-stage pipeline diagram showing semantic episode construction; SAM 3-style masks and SAM 3D-style metadata; frame-graph and temporal-track association; retrieval-ready memory writing; and option-blind retrieval followed by question answering.}
    \label{fig:pipeline}
\end{figure*}

Three requirements follow from this formulation. First, the same physical object should keep a stable identity across time. Second, object state should be described relative to stable references in the scene. Third, the video must be summarized into searchable segment-level units, because storing dense frame sequences would preserve redundancy rather than evidence and would make long-range retrieval expensive and brittle. Figure~\ref{fig:pipeline} summarizes the resulting end-to-end pipeline.

Figure~\ref{fig:pipeline} also clarifies that the method does not depend on a single heavy reconstruction stage. The semantic episode constructor narrows the local object inventory, the visual front-end extracts relative evidence within those windows, and the graph association stage decides persistence and anchor roles before memory writing. This staged decomposition is important for long egocentric video because errors remain local: a missed mask affects one episode, whereas failure of a global map would otherwise propagate across the whole day.

\subsection{Relative 4D scene graph representation}

At each sampled time step $t$, the method constructs a frame graph
\begin{equation}
G_t = (V_t, E_t),
\end{equation}
where $V_t = \{v_t^{space}, v_t^{act}\} \cup V_t^{obj}$ contains one space node, one activity node, and a set of object nodes. Each object node stores semantic and geometric attributes,
\begin{equation}
a_t(v) = [c_t, m_t, b^{3d}_t, x^{3d}_t, s_t],
\end{equation}
where $c_t$ is the semantic class, $m_t$ is the mask or crop reference inherited from the segmentation stage, $b^{3d}_t$ is a coarse 3D extent, $x^{3d}_t$ is a relative 3D location, and $s_t$ is a textual state descriptor. Edges encode relations such as \texttt{in}, \texttt{on}, \texttt{near}, and generic \texttt{spatial\_relation}, together with activity-linked interaction cues when available.

The representation becomes 4D after linking frame nodes into persistent tracks. Here, ``4D'' denotes a time-indexed relative 3D scene-graph memory, not a globally consistent 4D reconstruction. Let $p$ denote a persistent object track formed by associating semantically compatible observations across neighboring graphs. For each track, the method identifies whether it behaves as a \emph{static anchor} or a \emph{dynamic object}. Static anchors are objects whose semantic role and motion pattern are stable, such as tables, shelves, counters, or fridges. Dynamic objects are portable or manipulated items such as bags, cups, phones, tools, or food packages.

For a dynamic track $p$, the anchor-relative state at time $t$ is
\begin{equation}
z_t(p) = (a_t, d_t, u_t, s_t),
\end{equation}
where $a_t$ is the dominant nearby anchor, $d_t$ is a coarse relative distance, $u_t$ is a relative direction or layout cue, and $s_t$ is the current textual state. The representation is \emph{relative} because these quantities are defined with respect to stable anchors rather than to a global world frame.

A temporal event is written whenever a dynamic object's dominant anchor changes and the new anchor-relative state remains stable across time,
\begin{equation}
e_i = (p, [t_s,t_e], a^{-}, a^{+}, r_i),
\label{eq:temporal-event}
\end{equation}
where $[t_s,t_e]$ is the event span, $a^{-}$ and $a^{+}$ are the source and destination anchors, and $r_i$ is a short local rationale drawn from nearby action or interaction context. This event view explicitly records which object changed, when the change occurred, and which stable reference changed with it.

Table~\ref{tab:schema} summarizes the schema used in the representation and in the downstream memory.

\begin{table}[t]
\caption{Schema of R4DSG and its retrieval memory.}
\label{tab:schema}
\centering
\footnotesize
\begin{tabular}{@{}p{1.8cm}p{5.25cm}@{}}
\toprule
Component & Stored fields and role \\
\midrule
Frame node & Semantic class, mask or crop reference, relative 3D box, relative 3D location, textual state, and place association. Used to build $G_t$. \\
Frame edge & Subject, predicate, object, optional score, and relative spatial cue. Used to preserve relational structure within one frame graph. \\
Persistent track & Persistent ID, semantic class, temporal span, static or dynamic flag. Used to connect observations across time. \\
Temporal event & Persistent ID, time span, source anchor, destination anchor, local rationale. Used to write anchor-relative object change. \\
Retrieval-plus doc & Time span, place, activity, salient objects, object states, interaction summary, edge summary, lexical tokens, and optional explanation fields. Used for retrieval and QA. \\
\bottomrule
\end{tabular}
\end{table}

\begin{figure*}[t]
    \centering
  \includegraphics[width=0.96\textwidth]{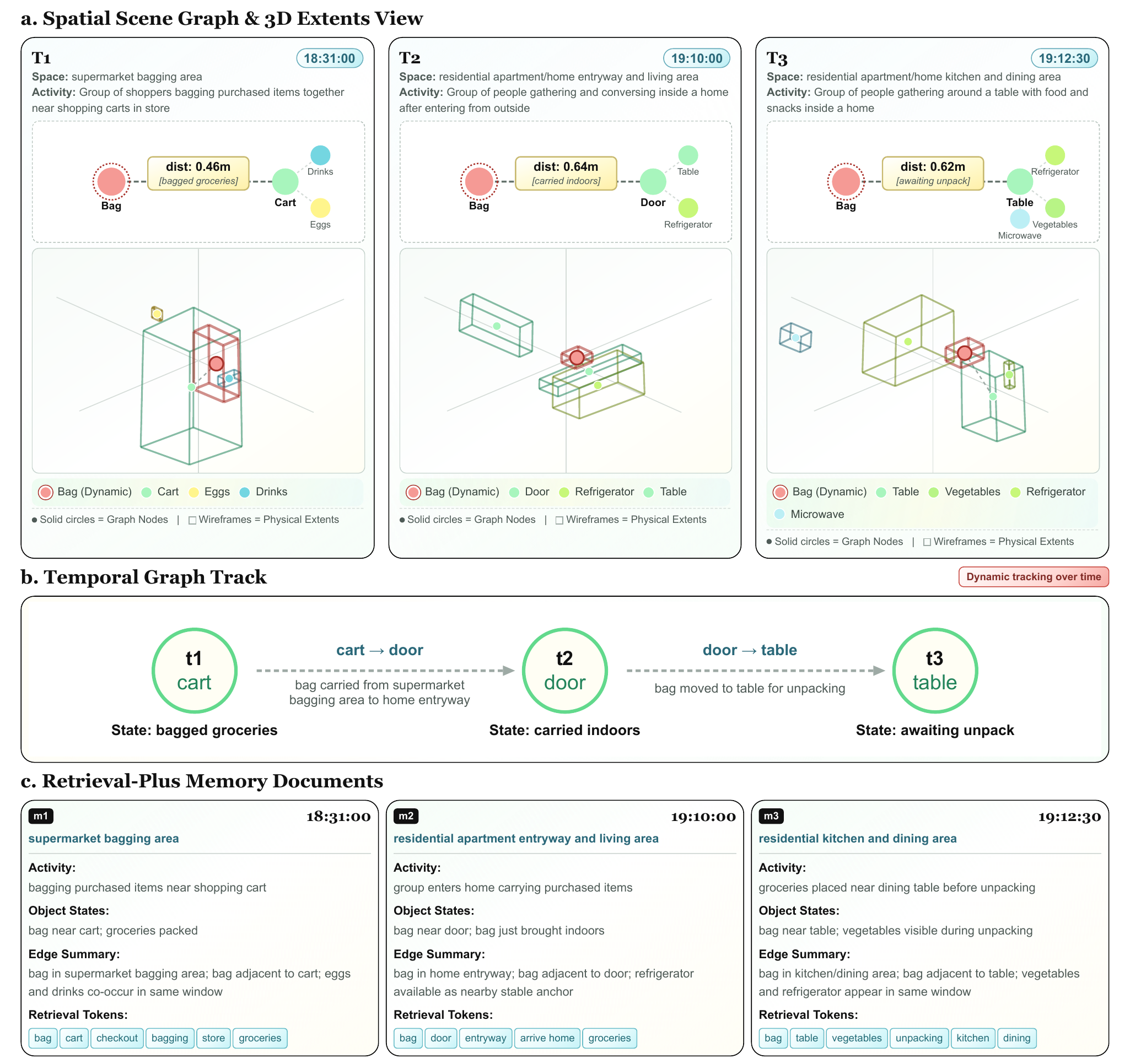}
    \caption{Bag-transfer example showing three anchor-relative states, the induced cart $\rightarrow$ door $\rightarrow$ table temporal track, and the resulting retrieval-plus memory documents. Top: scene graph and 3D extents at three timestamps; middle: the persistent track over anchors and state labels; bottom: the segment-level documents later retrieved for QA.}
    \Description{Three white-background panels visualize a grocery bag first near a cart, then near a door, and finally near a table, together with 3D extents and relations. A temporal graph below shows the cart-to-door and door-to-table transitions, and three memory cards summarize the corresponding retrieval-plus documents.}
    \label{fig:case4d}
\end{figure*}

Figure~\ref{fig:case4d} instantiates these variables on a concrete bag-transfer case selected from the released Day1-A1-JAKE scene-graph outputs: groceries are first bagged in a supermarket, then carried into the home entryway, and finally placed near the dining table for unpacking. This case was chosen because the anchor changes are unambiguous and the number of objects remains small enough for paper visualization. The three bands in Figure~\ref{fig:case4d} correspond to the same evidence at three abstraction levels: scene layout and coarse 3D extent, a persistent anchor-transition track, and the retrieval documents finally consumed by QA. The example therefore makes explicit that the memory documents are not detached summaries, but compact views written from the same anchor-relative state changes visualized in the graph.

\subsection{Queryable memory on top of the 4D graph}
\label{sec:queryable-memory}

Frame graphs and temporal events are still too local and too numerous to be queried directly over a day-scale video. We therefore write them into a retrieval-ready episodic memory. For an activity-centered segment window $S_k$, the corresponding memory entry is represented as
\begin{equation}
m_k = (\tau_k, \ell_k, \alpha_k, \mathcal{O}_k, \mathcal{Z}_k, \mathcal{R}_k, \mathcal{I}_k, \mathcal{T}_k, \mathcal{Y}_k),
\end{equation}
where $\tau_k$ is the time span, $\ell_k$ is the place, $\alpha_k$ is the activity, $\mathcal{O}_k$ is the salient object set, $\mathcal{Z}_k$ stores anchor-relative object states, $\mathcal{R}_k$ stores edge and relation summaries, $\mathcal{I}_k$ stores interaction or lexical-variant cues, $\mathcal{T}_k$ stores retrieval tokens, and $\mathcal{Y}_k$ stores optional explanation-oriented fields for \texttt{why} questions.

The document is written by the deterministic memory-writing operator $h$, which aggregates local frame graphs and overlapping temporal events,
\begin{equation}
m_k = h(\{G_t\}_{t \in S_k}, \{e_i\}_{e_i \cap S_k \neq \emptyset}).
\label{eq:memory-writer}
\end{equation}
Object nodes contribute class labels, object states, and relative 3D metadata; anchor-linked events contribute temporally compressed state transitions; and frame edges contribute the local relations that make the memory searchable by context rather than by object names alone. The memory is therefore written at the segment-window level, which keeps it compact while preserving enough relational evidence for QA.

This design directly supports downstream retrieval: \texttt{where} questions match place and anchor relations; \texttt{when} questions match time spans and anchor changes; object-state questions match salient objects, states, and edge summaries; and \texttt{why} questions can additionally use explanation fields. Because each entry can be serialized either as normalized text or as structured fields, the memory works with sparse, dense, or hybrid retrieval.

\subsection{End-to-end pipeline}

\subsubsection{Semantic episode construction}

A full egocentric day is too long, too redundant, and too semantically heterogeneous to be processed as one undifferentiated sequence. Before any object-centric reasoning happens, we therefore reorganize the raw video into semantic episodes. The video is segmented along two complementary axes: activity-oriented episodes and location-oriented episodes. Activity segments capture \emph{what is being done}, whereas location segments capture \emph{where it is taking place}. These two partitions are not redundant. Location windows stabilize anchor discovery, while activity windows provide the finer temporal units in which object states, interactions, and memory documents are written.

Each released segment record stores a timestamp span, a semantic label, and a controlled object vocabulary. We denote the resulting segment schema by
\begin{equation}
\sigma_k = (\tau_k, \ell_k, \alpha_k, C_k),
\end{equation}
where $\tau_k$ is the segment span, $\ell_k$ is the location label, $\alpha_k$ is the activity label, and $C_k$ is the controlled object set. These controlled objects become the semantic inventory for later segmentation, so the episode constructor already provides the scaffold that tells the front-end which frames belong together and which object categories are likely to matter.

\subsubsection{2D and 3D visual front-end with SAM 3-style consistency}

Once semantic episodes are available, the system first seeks a stable 2D object substrate inside each episode. We use a SAM 3-style promptable video segmentation stage~\cite{carion2025sam} followed by a SAM 3D-style RGB-only lifting stage~\cite{chen2025sam}. The input prompts for SAM 3 are not drawn from a fixed detector vocabulary. Instead, they come from the segment semantics described above: the activity and location labels specify the local context, and the controlled object set $C_k$ provides the target object inventory. In practice, SAM 3 is executed separately on activity-conditioned and location-conditioned windows so that the open-vocabulary search space stays bounded by episode semantics instead of ranging over the entire day.

This design is important for realistic egocentric video. Everyday scenes are open-vocabulary and highly variable, and the camera is uncalibrated and constantly moving. Segment-conditioned prompting gives SAM 3 local semantic context, while temporal mask propagation keeps the same prompted object visually consistent over neighboring frames. The output is therefore an episode-scoped bundle of prompted objects, masks, crops, and short local tracklets.

SAM 3D then lifts those observations into relative spatial metadata. The goal is not dense reconstruction, but the minimum 3D evidence needed for stable object-centric reasoning: \texttt{location\_3d}, \texttt{bounding\_box\_3d}, and pairwise spatial relations. These metadata provide coarse object size, relative object position, and object-anchor geometry even when no global world frame is available.

\subsubsection{Frame graph construction and persistent association}

The 3D metadata are then compiled into frame-wise scene graphs. Each graph contains one space node from the location label, one activity node from the activity label, and a set of object nodes from the lifted object metadata. Edges connect objects to spaces and to one another through containment, support, adjacency, and generic relative spatial predicates. In the why-aware extension, activity nodes may also carry explanation-bearing attributes such as \texttt{reason}, \texttt{purpose}, or \texttt{why\_summary}. This graph is the first fully symbolic interface in the pipeline: it converts visual evidence into a structure that can be aligned, compared, and accumulated over time.

Persistent identity is established in the next step. Since SAM 3 only guarantees local consistency inside an episode, the memory writer links graph nodes across neighboring frames using semantic compatibility, coarse 3D size continuity, relative 3D position continuity, and neighborhood consistency with nearby anchors. The association policy is deliberately conservative: if two observations disagree strongly, the track is split rather than force-merged; if a frame is missing an object because of blur or occlusion, the system keeps the neighboring track evidence instead of hallucinating a new state. Static anchors are inferred from objects that recur with low motion inside location-consistent windows, while dynamic objects are those whose dominant anchor or relative state changes over time. Because matching is performed in anchor-relative coordinates, the method never requires a globally aligned scene frame.

\subsubsection{Memory writing}

Once persistent tracks are available, the system writes two coupled memory views. The first is a temporal event stream, where a new event is emitted according to Eq.~\ref{eq:temporal-event}. The second is the retrieval-plus memory described in Sec.~\ref{sec:queryable-memory}, written at the segment-window level. This second view aggregates the graph evidence needed for QA: place, activity, salient objects, anchor-relative object states, interaction summaries, edge summaries, lexical variants, and temporal span. When why-aware upstream annotations are available, dedicated explanation fields are written into $\mathcal{Y}_k$.

This writing step is where the relative 4D representation becomes operational for retrieval. The memory does not attempt to replay the whole video, but it also does not collapse everything into a few abstract events. Instead, it preserves exactly the intermediate granularity needed by long-horizon retrieval: enough compression to search efficiently, enough relational detail to recover the correct object-state evidence once a question arrives.

\subsubsection{Question answering over retrieval-plus memory}

At inference time, a question is mapped to a retrieval query over the memory entries rather than over raw frames. The main branch retrieves the top relevant retrieval-plus documents and linearizes their structured fields into evidence text. In the conservative option-blind retrieval protocol, answer options are introduced only after retrieval, in the final multiple-choice prompt to the answer model. \texttt{Why} questions can optionally route to the explanation-aware memory view when those fields are available. The experiments in Sec.~\ref{sec:experiments} show that when the relevant object-state or explanation evidence has already been written into memory, retrieval-based QA becomes substantially more reliable.

\noindent \textbf{Reproducibility.} For each semantic episode, the pipeline (1) samples frames within the released activity/location windows, (2) constructs prompts from the episode labels and controlled object set, (3) lifts and associates observations using semantic, relative-size, relative-position, and anchor-neighborhood consistency, (4) emits persistent anchor changes using Eq.~\ref{eq:temporal-event}, and (5) writes segment documents using Eq.~\ref{eq:memory-writer}. QA retrieves the top eight documents and then applies the final multiple-choice prompt.

\section{Experiments}
\label{sec:experiments}

\subsection{Experimental settings}

\noindent \textbf{Dataset and evaluation.} We evaluate on the publicly available EgoLifeQA A1\_JAKE single-subject QA file, which contains 500 four-choice questions over seven recording days. Object-centric non-\texttt{who} filtering yields 255 questions, including 72 \texttt{when}, 15 \texttt{why}, and 149 with accessible target-time metadata. Accuracy is the primary metric. Our empirical claims are limited to this public single-subject split and do not imply multi-subject generalization.

\noindent \textbf{Implementation and scale.} All pipelines use Qwen3.5-27B as the answer model. R4DSG scene parsing from Sec.~\ref{sec:method} is held fixed during QA, and the main branch retrieves the top $k=8$ documents. The Day1-A1-JAKE stream used for all quantitative and qualitative experiments contains 828 clips and 06:51:50.52 of effective footage across five sessions.

\noindent \textbf{Baselines.} Beyond Plain RAG, EgoRAG-Text~\cite{yang2025egolife}, and VLM-only, we evaluate adapted EMQA-style episodic~\cite{datta2022episodic} and AMEGO-inspired active memories~\cite{goletto2024amego}, plus a No-Transition Object-Relation control. The adapted baselines are not official reproductions. The control uses the same cached visual evidence and local relations as R4DSG but removes persistent cross-segment identity and anchor-transition writing.

\subsection{Experimental Results}

We report overall and \texttt{when} accuracy under the question-only retrieval and conservative option-blind retrieval protocols, an exploratory \texttt{why} comparison, a memory-granularity ablation, and a compact memory-cost profile.

\subsubsection{Object-centric and temporal QA}

Table~\ref{tab:qa_controls} consolidates the main results. Under question-only retrieval, R4DSG improves over EgoRAG-Text by 6.7 points overall and 12.5 points on \texttt{when}. Under option-blind retrieval / no-why, it remains strongest, exceeding the AMEGO-inspired baseline by 1.2 and 5.6 points, respectively. The No-Transition control is lower than R4DSG despite sharing cached visual evidence and local relations, which is consistent with a benefit from persistent identity and anchor-transition memory rather than object/relation serialization alone.

\begin{table}[t]
\caption{Accuracy (\%): Overall uses all 255 questions and When uses the 72-question \texttt{when} subset. ``Question-only retrieval'' uses only the question to retrieve evidence; ``option-blind retrieval'' withholds answer options until final answer selection; ``no-why'' removes optional WhyMemory fields, not \texttt{why} questions.}
\label{tab:qa_controls}
\centering
\scriptsize
\setlength{\tabcolsep}{2.5pt}
\begin{tabular}{@{}p{2.35cm}p{2.8cm}cc@{}}
\toprule
Method & Protocol & Overall & When \\
\midrule
Plain RAG & question-only retrieval & 29.4 & 27.8 \\
EgoRAG-Text & question-only retrieval & 32.9 & 30.6 \\
VLM-only & question-only retrieval & 29.8 & 20.8 \\
R4DSG Retrieval+ & question-only retrieval & \best{39.6} & \best{43.1} \\
\midrule
EMQA-style episodic & option-blind retrieval & 32.2 & 30.6 \\
AMEGO-inspired active & option-blind retrieval & 36.1 & 37.5 \\
No-Transition obj.-rel. & option-blind retrieval & 34.9 & 34.7 \\
R4DSG Retrieval+ & option-blind retrieval / \mbox{no-why} & \best{37.3} & \best{43.1} \\
\bottomrule
\end{tabular}
\end{table}

\subsubsection{Explanation-sensitive evaluation with why-oriented memory}

Compared with Retrieval+, WhyMemory adds four explanation fields: \texttt{reason}, \texttt{purpose}, \texttt{utility}, and \texttt{why\_summary}. They are generated from local segment/activity context before QA, without ground-truth answers or answer options. Table~\ref{tab:why_extension} shows an increase from 33.3 to 40.0 on the 15-question \texttt{why} subset and a small overall lift. This result is exploratory because the subset is small; the no-why conservative result is reported separately in Table~\ref{tab:qa_controls}.

\begin{table}[t]
\caption{Exploratory explanation-oriented memory result (\%). Overall uses all 255 questions; Why uses the 15-question \texttt{why} subset.}
\label{tab:why_extension}
\centering
\footnotesize
\begin{tabular}{lcc}
\toprule
Metric & Retrieval+ & WhyMemory \\
\midrule
Overall & 39.6 & \best{40.0} \\
Why & 33.3 & \best{40.0} \\
\bottomrule
\end{tabular}
\end{table}

\subsubsection{Memory representation evaluation across compression granularities}

Event-only uses anchor-change events, Low-compression retains denser segment records, Episodic-only uses episode-level summaries, Hybrid combines event and episodic memories, and Retrieval+ uses the segment-window documents in Sec.~\ref{sec:queryable-memory}.

Table~\ref{tab:ablation_memory} shows that Retrieval+ has the highest observed overall and \texttt{when} accuracy. The lower observed accuracies of the alternative memory views are consistent with insufficient local context or a diluted change signal; this comparison does not isolate the causal mechanism. Retrieval+ retains the transition together with the place, activity, object, and interaction cues used for retrieval.

\begin{table}[t]
\caption{Memory-granularity ablation (\%). Overall uses all 255 questions; When uses the 72-question \texttt{when} subset.}
\label{tab:ablation_memory}
\centering
\footnotesize
\setlength{\tabcolsep}{3.8pt}
\begin{tabular}{lccccc}
\toprule
Metric & \makecell{Event-\\only} & \makecell{Low-\\compression} & \makecell{Episodic-\\only} & Hybrid & Retrieval+ \\
\midrule
Overall & 33.7 & 34.5 & 35.7 & 34.1 & \best{39.6} \\
When & 27.8 & 34.7 & 34.7 & 26.4 & \best{43.1} \\
\bottomrule
\end{tabular}
\end{table}

\subsubsection{Memory scale}

Table~\ref{tab:memory_cost} summarizes the offline memory scale. The 828 clips produce 2,476 frame-info entries and 134 Retrieval+ documents in a 0.58-MB JSON memory.

\begin{table}[t]
\caption{Offline memory scale profile.}
\label{tab:memory_cost}
\centering
\footnotesize
\begin{tabular}{lr}
\toprule
Measure & Value \\
\midrule
Clips / frame-info entries & 828 / 2,476 \\
Activity / location episodes & 178 / 114 \\
Retrieval+ documents / JSON size & 134 / 0.58 MB \\
Average whitespace tokens per document & 158.5 \\
\bottomrule
\end{tabular}
\end{table}

\section{Discussion and Limitations}

\noindent \textbf{Contribution and scope.} R4DSG changes the memory rather than the answer model: SAM3-style consistency and RGB-only lifting supply visual evidence, while relative graphs and Retrieval+ organize it for long-range QA. The clearest gain is on \texttt{when}; evidence is limited to the public A1\_JAKE single-subject split.

\noindent \textbf{Limitations and failure modes.} Errors may come from missed episodes; fragmented or duplicated tracks; cluttered anchors; retrieval misses; or insufficient causal or social context. Outputs are not yet a human-verified perception benchmark. Memory is retrospective and offline with limited cross-day persistence; online updates, stronger identity maintenance, and multi-subject validation remain future work.

\section{Conclusion}

R4DSG reframes long egocentric QA as memory construction over persistent objects, static anchors, and retrieval-ready documents. On object-centric EgoLifeQA, gains on \texttt{when} questions and the exploratory why-oriented extension suggest the value of temporal and explanatory memory. Relative scene graphs are a practical substrate for wearable assistants and embodied multimedia agents.

\begin{acks}
\begingroup\emergencystretch=2em
This work was supported by the New Generation Artificial Intelligence-\allowbreak National Science and Technology Major Project (Grant No. 2025ZD0122801) and The National Natural Science Foundation of China (Grant Nos. 62276063 and U23B2057).\par
\endgroup
\end{acks}

\bibliographystyle{ACM-Reference-Format}
\interlinepenalty=10000
\bibliography{software-2}
\vspace*{4\baselineskip}

\end{document}